\documentclass[10pt,twocolumn,letterpaper]{article}

\usepackage{arxiv}
\usepackage{times}
\usepackage{epsfig}
\usepackage{graphicx}
\usepackage{amsmath}
\usepackage{amssymb}

\usepackage{eqnarray}

\usepackage{subfigure}
\usepackage{epstopdf}
\usepackage{multirow}
\usepackage{color}
\usepackage{xcolor}
\usepackage{tabularx}
\usepackage{booktabs}
\usepackage{caption}
\usepackage{cite}
\usepackage{pifont}
\newcolumntype{L}[1]{>{\raggedright\arraybackslash}p{#1}}
\newcolumntype{C}[1]{>{\centering\arraybackslash}p{#1}}
\newcolumntype{R}[1]{>{\raggedleft\arraybackslash}p{#1}}
\newcommand*{\affaddr}[1]{#1} 
\newcommand*{\affmark}[1][*]{\textsuperscript{#1}}
\newcommand*{\email}[1]{\texttt{#1}}
 \usepackage{microtype}

\usepackage[pagebackref=true,breaklinks=true,letterpaper=true,colorlinks,bookmarks=false]{hyperref}

\finalcopy 

\ifarxivfinal\fi

\begin{document}

\title{FDAN: Flow-guided Deformable Alignment Network for Video Super-Resolution}

\author{%
Jiayi Lin\affmark[1,2], Yan Huang\affmark[1], Liang Wang\affmark[1]\\
\affaddr{\affmark[1]Center for Research on Intelligent Perception and Computing (CRIPAC),\\
National Laboratory of Pattern Recognition (NLPR),\\
Institute of Automation, Chinese Academy of Sciences (CASIA)}\\
\affaddr{\affmark[2]University of Chinese Academy of Sciences (UCAS)}\\
\email{\tt\small linjiayi18@mails.ucas.ac.cn}
\hspace{0.2cm}
\email{\tt\small \{yhuang, wangliang\}@nlpr.ia.ac.cn}
}
\maketitle
\ifarxivfinal\thispagestyle{empty}\fi

\begin{abstract}

Most Video Super-Resolution (VSR) methods enhance a video reference frame by aligning its neighboring frames and mining information on these frames. Recently, deformable alignment has drawn extensive attention in VSR community for its remarkable performance, which can adaptively align neighboring frames with the reference one. However, we experimentally find that deformable alignment methods still suffer from fast motion due to locally loss-driven offset prediction and lack explicit motion constraints. Hence, we propose a Matching-based Flow Estimation (MFE) module to conduct global semantic feature matching and estimate optical flow as coarse offset for each location. And a Flow-guided Deformable Module (FDM) is proposed to integrate optical flow into deformable convolution. The FDM uses the optical flow to warp the neighboring frames at first. And then, the warped neighboring frames and the reference one are used to predict a set of fine offsets for each coarse offset. In general, we propose an end-to-end deep network called Flow-guided Deformable Alignment Network (FDAN), which reaches the state-of-the-art performance on two benchmark datasets while is still competitive in computation and memory consumption.

\end{abstract}

\section{Introduction}

	Video Super-Resolution (VSR) is the task of increasing the resolution of video frames, which is widely applied in video surveillance, satellite imagery, etc. One of the main challenges of VSR comes from the misalignment between neighboring low-resolution (LR) frames and the reference one, which causes underuse of temporal compensatory information and even leads to artifacts. Therefore, to utilize neighboring frames properly, alignment is of considerable importance.
	\begin{figure}[t] 
	\setlength{\abovecaptionskip}{0.cm}
	\setlength{\belowcaptionskip}{-.5cm}

	\centering

	 \subfigure[ref frame]			{\label{fig:Alignment_1}\includegraphics[scale=.36]{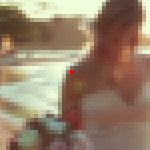}}\hspace{.085cm}
	 \subfigure[flow-based]			{\label{fig:Alignment_2}\includegraphics[scale=.36]{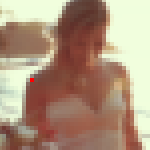}}\hspace{.085cm}
	 \subfigure[deformable]			{\label{fig:Alignment_3}\includegraphics[scale=.36]{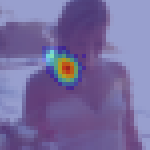}}
	 \subfigure[FDA (proposed)]		{\hspace{.1cm}\label{fig:Alignment_4}\includegraphics[scale=.36]{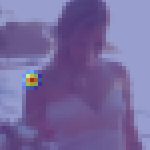}\hspace{.2cm}}
	
	\caption{Visual comparison of the weighted sampling on the neighboring frame using different alignment methods. (a) is the reference frame and the red point denotes the location to compensate. (b) and (d) is given by our proposed method, while (c) applies the alignment module PCD in EDVR~\cite{EDVR}. } 
	\label{fig:vis_dcnv}
	\end{figure}

	Recent methods on frame alignment can be roughly classified into two categories: flow-based alignment~\cite{TOFlow-Vimeo,RBPN} and deformable alignment~\cite{TDAN,EDVR}. The former estimates the optical flow and warps neighboring frames to align with the reference one. While the latter performs deformable convolution~\cite{DeformableCN,DeformableCNv2} to align neighboring frames, which is more lately proposed and has achieved remarkable performance~\cite{EDVR}. Both methods can be seen as performing sampling on neighboring frames to compensate the reference one. The difference is that, to compensate one spatial location in a reference frame (Fig.~\ref{fig:Alignment_1}), the former samples the feature from one location guided by flow (Fig.~\ref{fig:Alignment_2}), while the latter adaptively samples features from several locations guided by the predicted offsets that gives a set of sampling locations (Fig.~\ref{fig:Alignment_3}). Due to more diverse sampling, deformable alignment tends to perform better than flow-based alignment~\cite{UnderstandDcnv}.

	\begin{figure*}[t] 
	\setlength{\abovecaptionskip}{0.cm}
	\setlength{\belowcaptionskip}{-.5cm}
	\vspace{-.3cm}
	\centering

	 \subfigure[fast]		{\label{fig:offsetDistri_fast}\includegraphics[scale=.4]{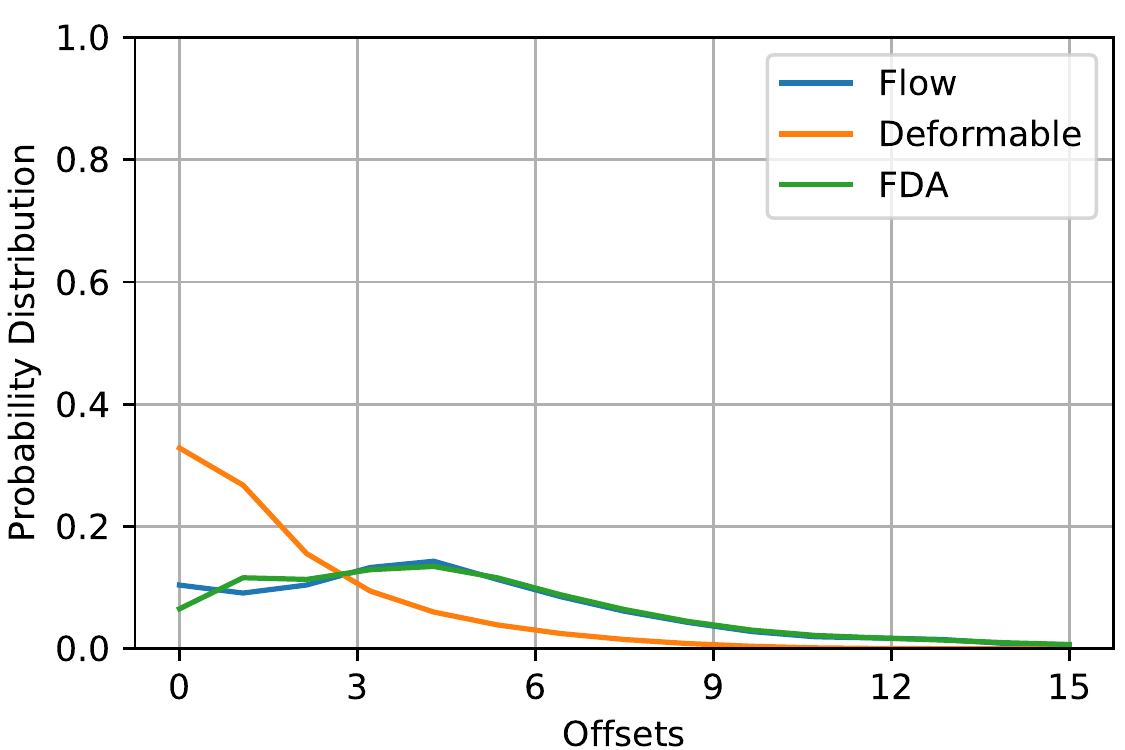}}\hspace{.5cm}
	 \subfigure[medium]		{\label{fig:offsetDistri_medium}\includegraphics[scale=.4]{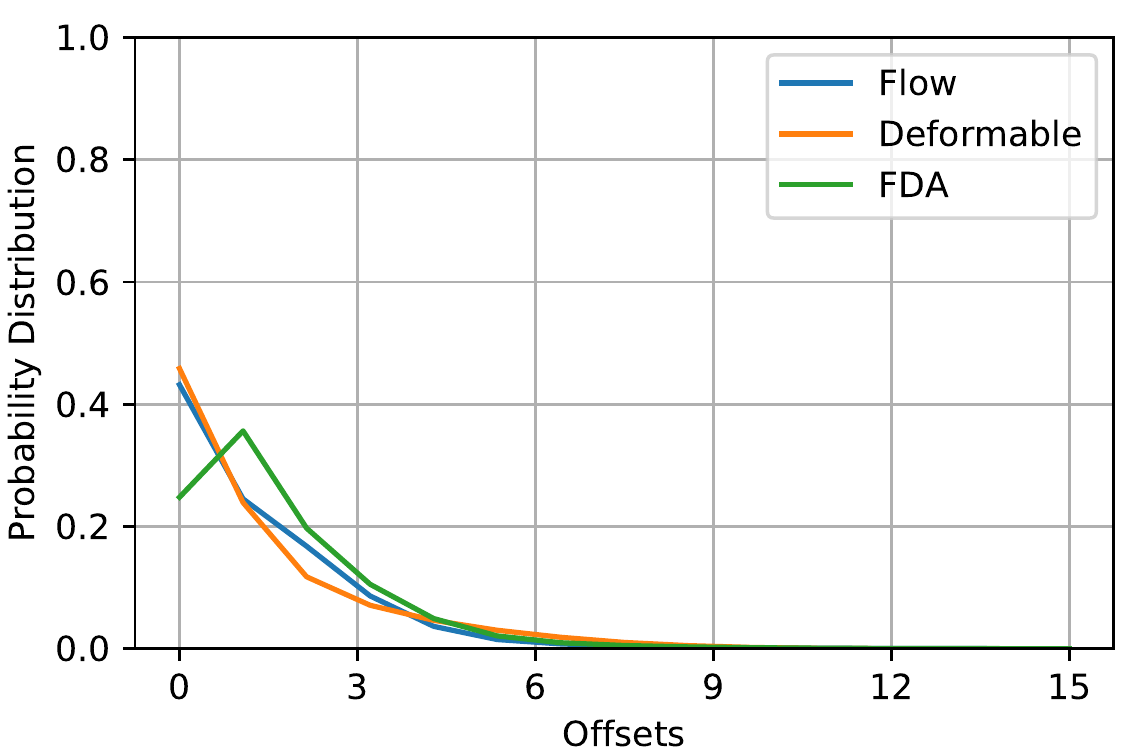}}\hspace{.5cm}
	 \subfigure[slow]		{\label{fig:offsetDistri_slow}\includegraphics[scale=.4]{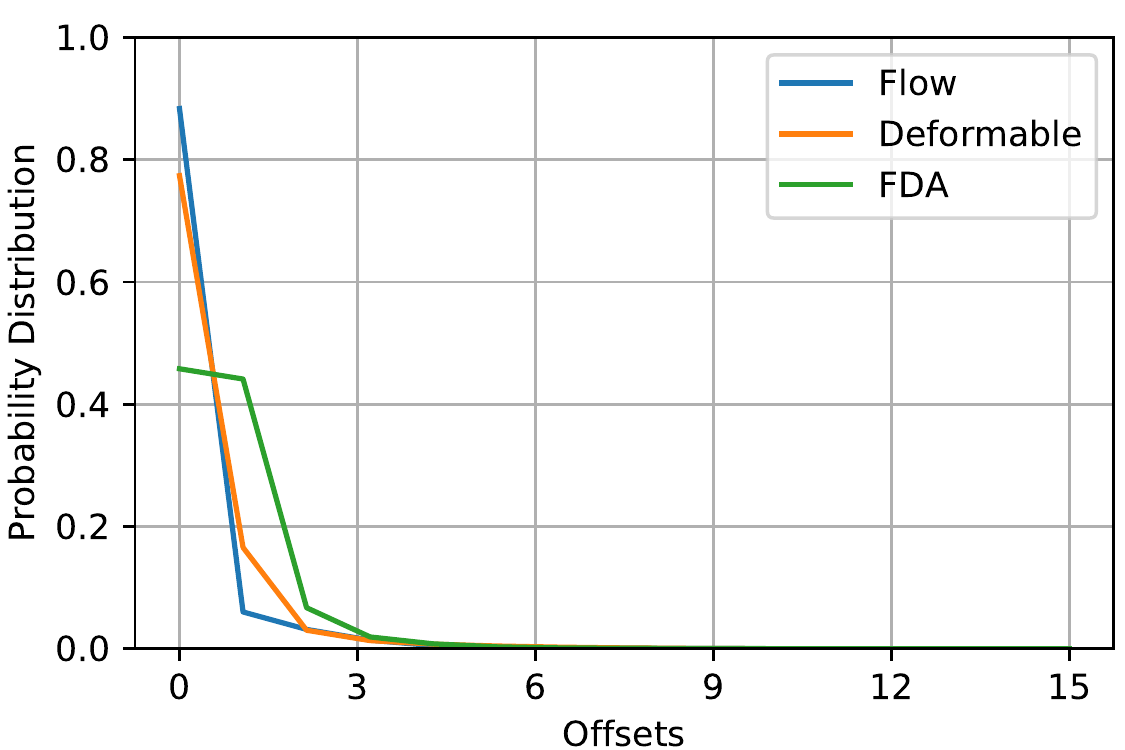}}

	\caption{Comparison of the distribution of the offset value using different alignment methods. We count the offsets that give sampling locations on frame $t-3$ to compensate the reference frame $t$. ``Flow'' is given by MFE in FDAN and ``Deformable'' is given by PCD in EDVR~\cite{EDVR}. The distributions are calculated on Vimeo90K-T~\cite{TOFlow-Vimeo} divided into fast, medium, slow motion respectively as in ~\cite{RBPN}. Note that the probability distribution of offsets larger than 15 is too small to be statistically significant and not shown here. Fractions are rounded down.}
	\label{fig:offsetDistri}
	\end{figure*}

	However, deformable alignment still suffers from fast motion. Since the sampling locations (offsets) of deformable convolution are predicted by a fully convolutional module (such as PCD~\cite{EDVR}), the experimental study shows that the
	empirical perceptive field is limited. 
	Thus, the offset prediction is relatively local, which can be verified by the phenomenon that the learned offsets are mostly small in value even on videos with fast motion,
	as shown in Fig.~\ref{fig:offsetDistri}. While the global-matching-based flow estimation methods~\cite{SFNet,FlowNet} can handle various motion more effectively, especially fast motion.

Therefore, we propose an alignment method call Flow-guided Deformable Alignment (FDA) to help the learned offsets more precise, which is a complementary integration of flow-based alignment and deformable alignment (Fig.~\ref{fig:Alignment_4}). Specifically, FDA is implemented by two modules: the Matching-based Flow Estimation module (MFE) and the Flow-guided Deformable Module (FDM). Rather than performing the flow estimation and then deformable convolution sequentially, our MFE module globally estimates the optical flow as coarse offset to help the prediction of a set of fine offsets in deformable convolution. In our FDM, by applying our proposed Flow-guided Deformable Convolution (FDC), we have the sampling locations near the true correspondence and mainly on the same object so as to suppress irrelevant sampling noises. In this way, most learned offsets of deformable convolution is around the center location and our method is able to handle fast motion (Fig.~\ref{fig:offsetDistri}).

Moreover, the flow estimation module MFE is designed to achieve less consumption. Specifically, MFE applies an all-pairs matching strategy to capture motion globally, which may cause huge computation and memory consumption. We perform the matching at $1/4$ resolution to get the coarse flow first and then upsample it, which is not a bottleneck in computation or memory even when the output size is 4K. Besides, we need no pretraining with ground truth flow or finetuning on the degraded datasets as in ~\cite{TOFlow-Vimeo}. Certainly, MFE can be replaced with any flow estimation module for better performance, we only try to reach a balance between the performance and efficiency here and use it to illustrate the effectiveness of explicit motion constraints.

Based on FDA, we propose Flow-guided Deformable Alignment Network (FDAN), an end-to-end deep network architecture for robust alignment epsecially in VSR. Furthermore, our FDAN follows a concise cascading architecture with strong expandability.
The contributions of this paper are three-fold. (1) We propose a novel alignment method call Flow-guided Deformable Alignment (FDA), which is able to handle fast motion by introducing optical flow to deformable alignment. (2) We propose a Matching-based Flow Estimation module (MFE) to estimate optical flow globally, and plug it into an end-to-end trainable network without pretraining or other supervision in the loss function. (3) we propose Flow-guided Deformable Alignment Network (FDAN) based on FDA, which achieves the state-of-the-art performance on Vimeo90K-T and UDM10 and is still competitive in computation and memory consumption.

\section{Related Work}
The first deep learning based method on Single Image Super-Resolution (SISR), which has driven rapid development of SISR as well as VSR.
The key difference between SISR and VSR is that VSR can benefits from temporal compensatory information in neighboring frames. The typical architecture of VSR model consists of 4 components: a feature extractor, an alignment module, a temporal fusion module and a reconstruction module. Since nearly any design in SISR can be readily integrated into the feature extractor and the reconstruction module, the task of VSR usually focuses on the alignment module and the temporal fusion module. In this work, we concentrate on the design of the alignment module, which can be roughly divided into two categories, the implicit alignment and the explicit one.

{\bf Implicit Alignment.} Recently, several methods attempt to side-step explicit alignment between frames by directly adopting sequential modules or models like 3D convolution~\cite{DUF, FSTRN, TGA} or Recurrent Neural Network~\cite{BRCN, RSDN} in the further fusion step. They have achieved considerable success and keep the architecture elegant. However, they are insufficient to deal with fast motion, which introducing undesired information and thus worsening the following fusion process. Therefore, explicit alignment should be still unavoidable in order to handle fast motion.

	\begin{figure*}[ht] 
	\setlength{\abovecaptionskip}{-.1cm}
	\setlength{\belowcaptionskip}{-.4cm}
	\begin{center}
	\includegraphics[scale=.9]{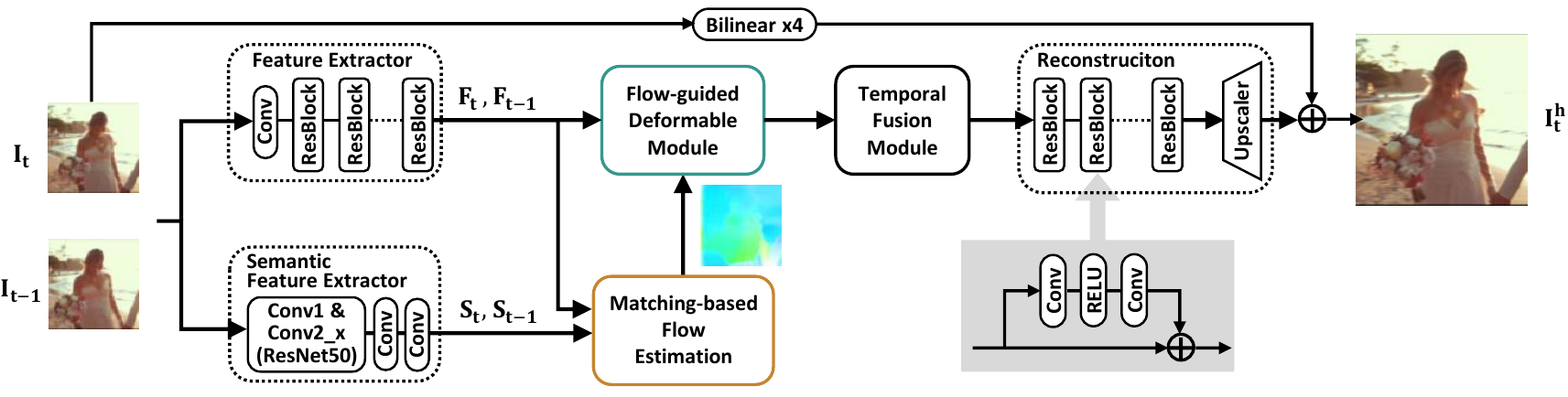}
	\end{center}
	\caption{The proposed FDAN framework. The top branch is for restoration of the reference LR frame, and the bottom one is used to provide the optical flow. The alignment method FDA is implemented the Matching-based Flow Estimation module (MFE) and the Flow-guided Deformable Module (FDM). We only show one neighboring frame here as an example.
	}
	\label{fig:2}
	\end{figure*}

{\bf Explicit Alignment.} By directly adopting the pre-computed optical flow~\cite{VSRnet,RBPN} or plugging an optical flow estimation module into an end-to-end model~\cite{VESPCN,TOFlow-Vimeo}, the motion is appropriately captured thus beneficial to the following process. TOFlow~\cite{TOFlow-Vimeo} suggests that using a plugged-in alignment module can be better than using an external flow estimation network, since the flow is task-oriented. However, directly generating flow may not be the best choice as there is no ground truth flow for supervision, making the network difficult to learn. In other words, the flow is generated in a regression way that minimizes pixel-level difference of the output SR image and its ground truth, so it might not generalize well in a new scene.

For better generalization and robustness, the matching-based method rises recently. TTSR~\cite{TTSR} uses a transformer~\cite{Vaswani2017AttentionIA} to query a similar feature to compensate its target feature, while MuCAN~\cite{MuCAN} adopts KNN strategy to find $k$ similar features and aggregate them to provide information for their target feature. However, for efficiency, these methods always need to constraint a fixed searching window that may limit their ability capture fast motion. In this work, our MFE is a combination of matching-based and generating methods, trying to reach a balance between performance and efficiency.

Furthermore, deformable convolution~\cite{DeformableCN}, usually used for high-level vision tasks such as object detection ~\cite{STSN}, is creatively adopted in VSR, increasing sampling diversity and boosting the performance of VSR. By weighting different sample features differently, it is more flexible and weakens the influence of inappropriate sampling. TDAN~\cite{TDAN} adopts a cascade of 4 deformable convolutional layers to align neighboring frames to the reference one, while EDVR applies a pyramid architecture to capture fast motion. However, deformable convolution could lead to unstable training process and limited generalization. Our work uses explicit alignment to handle fast motion by creatively integrating flow to deformable convolution using our proposed FDC.

\section{Flow-guided Deformable Alignment Network}
\subsection{Overview}
	Given a $2N+1$ consecutive LR frame sequence $\lbrace I_{t-N},\cdots,I_{t},\cdots,I_{t+N}\rbrace$ with $I_{t}$ as the reference frame, our task is to restore the reference HR frame $I_{t}^h$. In general, FDAN follows the typical architecture of VSR model, consisting of four components: a feature extractor, an alignment module, a temporal fusion module and a reconstruction module, as shown in Fig.~\ref{fig:2}.

	In FDAN, the alignment is implemented by two modules: (1) Matching-based Flow Estimation module (MFE) that estimates the optical flow between frames, elaborated in Sec.~\ref{subsection:MFE}. (2) Flow-guided Deformable Module (FDM) that perform sampling on neighboring frames using a cascade of two deformable convolutional layers, elaborated in Sec.~\ref{subsection:dcnv}. To highlight the effectiveness of our alignment module, we keep the other three modules simple and regular, elaborated in Sec.~\ref{subsection:ImplDtl}.

\subsection{Matching-based Flow Estimation}\label{subsection:MFE}
\begin{figure}[t] 
	\setlength{\abovecaptionskip}{.16cm}
	\setlength{\belowcaptionskip}{-.5cm}
	\begin{center}
	\includegraphics[scale=0.7]{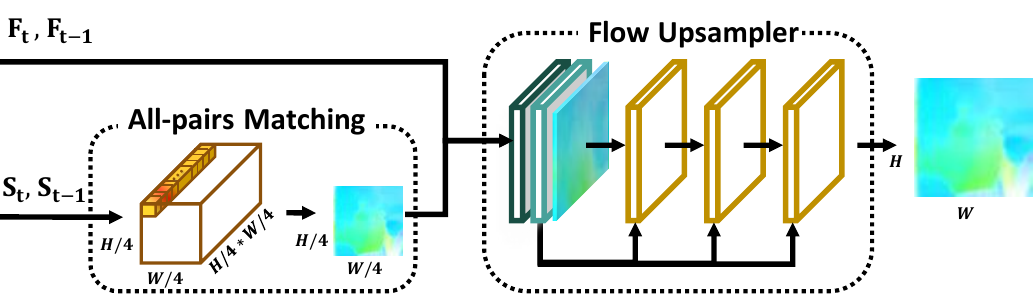}
	\vspace{-.35cm}
	\end{center}
	\caption{Matching-based Flow Estimation module (MFE). The flow is generated in a coarse-to-fine manner.}
	\label{fig:3}
	\end{figure}

	To estimate the optical flow in an end-to-end framework, previous methods~\cite{VESPCN, SPMC, TOFlow-Vimeo} usually use stacked convolutional layers to directly generate the flow like SpyNet~\cite{SpyNet}, which are difficult to learn the true flow or generalize well. While the matching-based methods~\cite{MuCAN} can alleviate this problem but usually introduces massive computation and memory consumption. Therefore, in our MFE (Fig.~\ref{fig:3}), with the prior that the optical flow is usually sparse, we perform all-pairs matching at $1/4$ resolution to get the coarse flow $O^{c}\in R^{H/4\times W/4\times 2}$ and later generate the fine one $ O^{f} \in R^{H\times W\times 2}$ to reach the computation and memory efficiency.

{\bf All-pairs Matching.}
	To get the features for matching, inspired by SFNet~\cite{SFNet} which use high-level semantics for the task of semantic correspondence, we propose Semantic Feature Extractor to adapt the matching task, which is made of the first two blocks of Resnet50~\cite{Resnet} referred to as ``conv1'' and ``conv2\_x'' followed by two convolutional layers. Later, the generated feature maps $\lbrace S_{t-N},\cdots,S_{t},\cdots ,S_{t+N}\rbrace$ are normalized and used to perform matching, where $S_t\in R^{H/4\times W/4\times C_{1}}$.

	Taking matching $S_{t-1}$ and $S_{t}$ as an example, the matching score between feature vectors such as $s_{t}^{i}\in R^{1\times 1\times C_{1}}$ from $S_{t}$ and $s_{t-1}^{j}$ from $S_{t-1}$ is calculated as the normalized inner product as in FlowNet~\cite{FlowNet}, where $i$ and $j$ denote the spatial location. To further infer the coarse flow $O^{c}$, instead of directly getting the coordinate of the feature with the highest matching score, we adopt the method in ~\cite{SFNet} to ensure this operation derivable by applying the Gaussian filter.

{\bf Flow Upsampler.}
	To generate the fine flow $O^f$ from the coarse one $O^c$, $\lbrace I_{t-N},\cdots ,I_{t},\cdots,I_{t+N}\rbrace$ are first encoded with the Feature Extractor to generate $\lbrace F_{t-N},\cdots,F_{t},\cdots ,F_{t+N}\rbrace $, where $F_t\in R^{H\times W\times C_{2}}$. At the same time, we interpolate $O^c$ by scale 4 with the nearest neighbor strategy to get the initial fine flow $O^{if}$ and use it to warp the neighboring feature map to align the reference one. For example, $F_{t-1}$ is warped to generate $ F_{t-1}^{if}$ with $O_{t-1}^{if}$.

	So far, we get $ F_{t}$, $ F_{t-1}^{if}$ and $O_{t-1}^{if}$. They are then concatenated together to generate the final fine flow $O_{t-1}^{f}$, which is implemented with a dense block as in FlowNet~\cite{FlowNet} but much lighter, since there is only sligt motion between $F_{t}$ and $F_{t-1}^{if}$ and we do not need deeper network to enlarge the perceptive field to capture fast motion. 

	In this way, the noises in the coarse flow is refined, while the fine flow generation module under the assistance of the coarse flow can generalize well and need fewer parameters, which reaches a balance between the performance and efficiency for flow estimation.

\subsection{Flow-guided Deformable Convolution} \label{subsection:dcnv}
\begin{figure}[t]
	\setlength{\abovecaptionskip}{0pt}
	\setlength{\belowcaptionskip}{-.8cm}
	  \centering
	  \subfigure[Naive Integration]{
	    \label{fig:subfig:4a} 
	    \includegraphics[scale=.85]{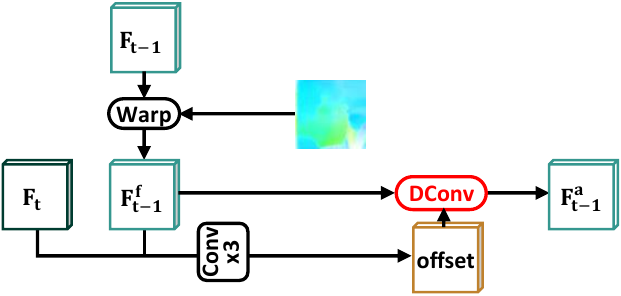}}
	  \hspace{1in}
	  \subfigure[Advanced Integration]{
	    \label{fig:subfig:4b} 
	    \includegraphics[scale=.85]{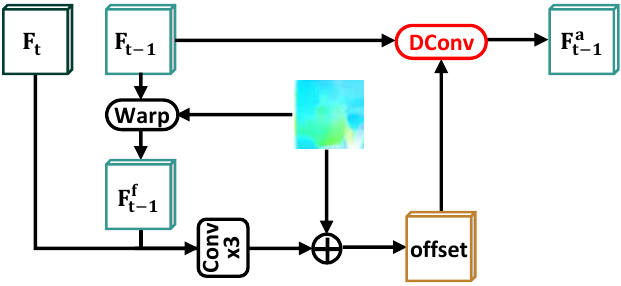}}
	  \caption{A comparison of two strategies for flow integration. Naive Integration directly uses flow to warp neighboring frames, and Advanced Integration uses flow to help generation of the offsets in deformable convolution.}
	  \label{fig:4} 
	\end{figure}

We apply a cascade of two deformable convolutional layers, in which we integrate the flow into the first one called Flow-guided Deformable Convolution (FDC) to deal with fast motion first~\ref{fig:4} and then keep the second one the same as in EDVR~\cite{EDVR} for further refinement.

Deformable convolution~\cite{DeformableCN} was first applied to VSR in TDAN~\cite{TDAN} to predict the sampling locations and was further extended to predict the weight of each sample using modulated deformable convolution~\cite{DeformableCNv2} in EDVR~\cite{EDVR}. We use modulated deformable convolution and call it deformable convolution for brevity.
In practice, taking $F_{t}$ and $F_{t-1}$ as an example, their concatenation in the channel direction $[F_{t}, F_{t-1}]$ is used to generate the offsets and the corresponding modulation scalars through several convolutional layers.

\begin{figure}[] 
	\setlength{\abovecaptionskip}{-.0cm}
	\setlength{\belowcaptionskip}{-.50cm}
	\centering
	 \subfigure[$F_{t}$]{\label{fig:subfig:inti1}\includegraphics[width=0.72in]{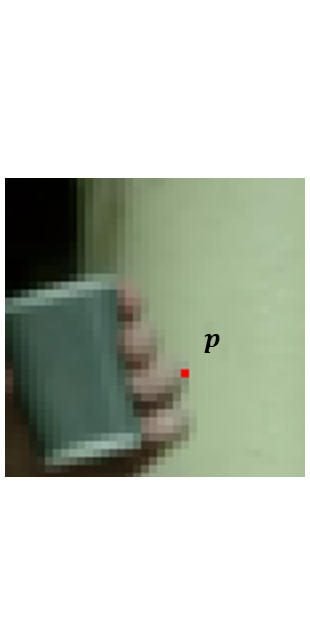}} \hspace{0.01cm}
	 \subfigure[$F_{t-1}$]{\label{fig:subfig:inti2}\includegraphics[width=0.72in]{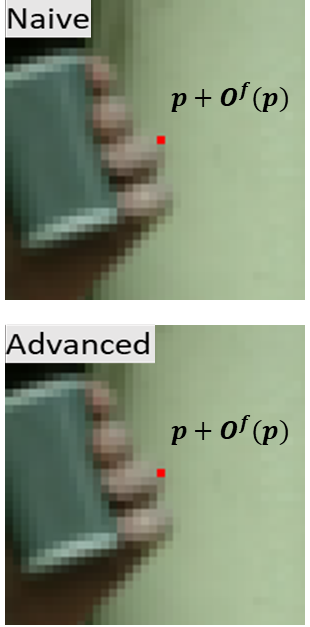}} \hspace{0.01cm}
	 \subfigure[$F_{t-1}^f$]{\label{fig:subfig:inti3}\includegraphics[width=0.72in]{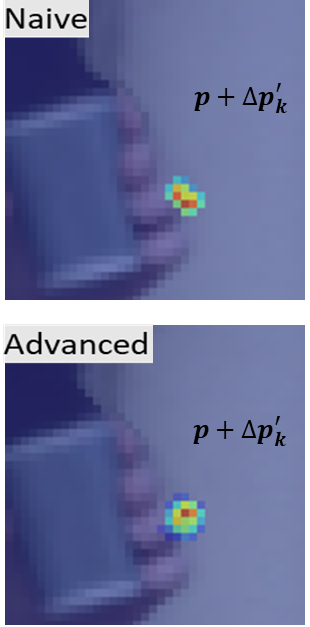}} \hspace{0.01cm}
	 \subfigure[$F_{t-1}$]{\label{fig:subfig:inti4}\includegraphics[width=0.72in]{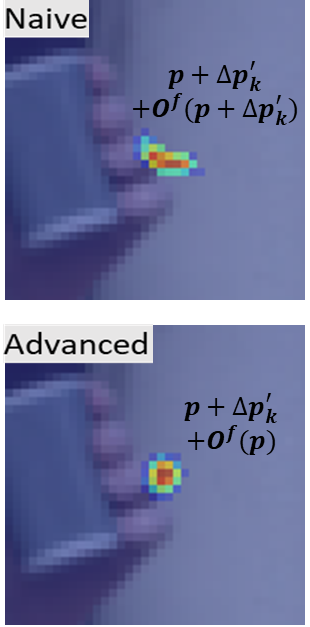}}	
	 \caption{A comparison of sampling on featrue maps using Naive Intigration and Advanced Intigration. (a) shows the location to compensate on the reference frame. (b) shows the corresponding location given by the flow $O^{f}$. (c) shows the sampling on $F_{t-1}^{f}$ given by $\lbrace\Delta p_{k}\rbrace_{k=1}^{K}$ and $\lbrace\Delta m_{k}\rbrace_{k=1}^{K}$. In (d), ``Naive'' shows corresponding samples on $F_{t-1}$ warped from (c), while the ``Advanced'' shows the sampling on $F_{t-1}$.}\label{fig:5}
	\end{figure}

In a deformable convolutional layer with a $3\times 3$ kernel, the offsets $\lbrace\Delta p_{k}\rbrace_{k=1}^{K}$ and the modulation scalars $\lbrace\Delta m_{k}\rbrace_{k=1}^{K}$ for location $p$ correspond to $p_{k}\in \lbrace(-1,-1),(-1,0),\cdots,(1,1)\rbrace$, where $K=9$. So the aligned features $F_{t-1}^{a}$ at location $p$ can be obtained by:
	\begin{equation}
		\setlength{\abovedisplayskip}{0.22cm}
		\setlength{\belowdisplayskip}{0.22cm}
		F_{t-1}^{a} (p)= \sum_{p_{k}}{w_{k}\cdot F_{t-1}( p+p_k+ \Delta p_{k} )\cdot \Delta m_k}, \label{eq1}
	\end{equation}
where $w_{k}$ denotes the weight of deformable convolution.
As shown in Fig.~\ref{fig:Alignment_3}, using deformable convolution alone without explicit motion estimation cannot find the true correspondences. Therefore, we use the flow $O^{f}$ to give true correspondences (Fig.~\ref{fig:subfig:inti2}) as the coarse offset, and explore how to generate a set of fine offsets to perform diverse sampling.
For brevity, we express the total offset from $p$ as:
	\begin{equation}
		\setlength{\abovedisplayskip}{0.22cm}
		\setlength{\belowdisplayskip}{0.22cm}
		\Delta p_{k}^{'}= p_k+\Delta p_{k}. \label{eq3}
	\end{equation}

To integrate the flow into deformable convolution, we first warp $F_{t-1}$ by $O^{f}$ to get $F_{t-1}^f$. So, the relation between $F_{t-1}$ and $F_{t-1}^f$ can be expressed as:
	\begin{equation}
		\setlength{\abovedisplayskip}{0.22cm}
		\setlength{\belowdisplayskip}{0.22cm}
		F_{t-1}^f (p)=F_{t-1} (p+O^{f} (p)).  \label{eq4}
	\end{equation}
Then $[F_{t}, F_{t-1}^f]$ is used to generate the offsets and the modulation scalars through $3$ convolutional layers.
In this way, we get $\lbrace\Delta p_{k}\rbrace_{k=1}^{K}$ and $\lbrace\Delta m_{k}\rbrace_{k=1}^{K}$ for location $p$ as shown in Fig.~\ref{fig:subfig:inti3} and use them to generate the aligned features $F_{t-1}^{a}(p)$.

{\bf Naive Integration (Fig.~\ref{fig:subfig:4a}).} An intuitive thought is to perform deformable convolution on the warped neighboring feature map $F_{t-1}^f$.
The aligned feature at $p$ is:
	\begin{equation}
		\setlength{\abovedisplayskip}{0.22cm}
		\setlength{\belowdisplayskip}{0.22cm}
		F_{t-1}^{a} (p)=\sum_{p_{k}}{w_{k}\cdot  F_{t-1}^{f}( p+\Delta p_{k}^{'} )\cdot \Delta m_k},  \label{eq4}
	\end{equation}
which corresponds to:
	\begin{equation}
		\setlength{\abovedisplayskip}{0.22cm}
		\setlength{\belowdisplayskip}{0.22cm}
		F_{t-1}^{a} (p) = \sum_{p_{k}}{w_{k}\cdot  F_{t-1}( p+\Delta p_{k}^{'} +O^{f} (p+\Delta p_{k}^{'}) )\cdot \Delta m_k}. \label{eq5}
	\end{equation}
However, as shown is Fig.~\ref{fig:subfig:inti4}, Naive Integration can introduce noises as spatial neighboring locations on $F_{t-1}^{f}$ may not be neighbors on $F_{t-1}$.
So we constraint all samples for $p$ to share one flow value $O^{f}(p)$, so that they are more likely to be neighbors and on one object.

{\bf Advanced Integration (Fig.~\ref{fig:subfig:4b}).} The aligned feature for location $p$ is modified as:
	\begin{equation}
		\setlength{\abovedisplayskip}{0.22cm}
		\setlength{\belowdisplayskip}{0.22cm}
		F_{t-1}^{a} (p)= \sum_{p_{k}}{w_{k}\cdot  F_{t-1}( p+\Delta p_{k}^{'} +O^{f} (p))\cdot \Delta m_k}.  \label{eq5}
	\end{equation}
 Advanced Integration directly performs sampling on the original feature map $F_{t-1}$.

Compared to Naive Integration, Advanced Integration requires that all the sampling locations to compensate location $p$ on $F_{t}$ to share the same flow value $O^{f} (p)$. Therefore, the sampling is performed around the true correspondence on $F_{t-1}$, rather than spatial neighboring locations on $F_{t-1}^{f}$ that might be irrelevant due to uncorrect flow estimation.


\section{Experiments}
\subsection{Implementation Details}\label{subsection:ImplDtl}
	{\bf Network details.}
	In FDAN, Feature Extractor consists of a convolutional layer and a cascade of 5 residual blocks. Temporal Fusion Module is completed with 3 convolutional layers in an attention way inspired by ~\cite{TGA,EDVR}. Reconstruction modules is a cascade of 10 residual blocks and Upscaler. The residual block is shown in Fig.~\ref{fig:2} which is widely used and can be replaced by any other advanced modules like residual dense block~\cite{RDN,DUF} or SD Block ~\cite{RSDN}. In Upscaler, inspired by the EDSR~\cite{EDSR}, we utilize two sub-pixel convolution~\cite{sub-pixel-convolution} followed by a convolutional layer for the final ×4 upscaling. All convolutional layers in feature extractors are followed by LeakyReLU~\cite{LeakyReLU}, except for those in the residual blocks. The dimension of feature map $C_{1}=128$, $C_{2}=128$. The number of input frames is $7$.

	{\bf Dataset.} We adopt Vimeo90K~\cite{TOFlow-Vimeo} as our training set, which is widely used for VSR~\cite{TDAN,TGA,EDVR}. We crop patches of size $256\times 256$ from high resolution video clips and use it to generate low-resolution patches of $64 \times 64$ by applying a Gaussian blur with a standard deviation of $\sigma = 1.6 $ and $4\times $ downsampling as in ~\cite{DUF,TOFlow-Vimeo,TGA}. Furthermore, we augment the training dataset by flipping and rotating with a probability of $0.5$. We evaluate our proposed method on 4 popular benchmarks: Vimeo90K-T, UDM10~\cite{PFNL-UDM10}, SPMCS~\cite{SPMC}(only for ablation study) and Vid4~\cite{Vid4}. Vimeo90K-T contains about 7.8k high-quality clips of 7 frames and various motion types, which can be roughly divided into fast, medium and slow motion~\cite{RBPN}. UDM10 and SPMCS consists of 10 and 30 videos of higher resolution ($720 \times 1272$ and $540 \times 960$ respectively) with various motion. Vid4 consists of 4 videos with relatively slow motion.

	{\bf Training details.} During training, the model is supervised by pixel-wise L1 loss. We use Adam ~\cite{Adam} optimizer where $\beta_{1} = 0.9$ and $\beta_{2} = 0.99$ to optimize our model. The learning rate is initially set to $1e-4$ and updated by CosineAnnealingLR in PyTorch with a period of $300$ epochs and the minimum learning rate is set to $1e-6$. Our training process is one-step containing 300 epochs with batch size of $16$.

	\begin{table*}[]
	\setlength{\abovecaptionskip}{-5pt}
	\setlength{\belowcaptionskip}{0pt}
	\caption{Quantitative comparison with other state-of-the-art VSR methods on Vimeo90K-T, Vid4 and UDM10 for $4\times$ VSR. * means the values are taken from ~\cite{RSDN}. Others are taken from their publications. \textcolor{red} {Red }text denotes the best performance and \textcolor{blue} {blue} text denotes the second best performance.}\label{table:1}
	\footnotesize
	\begin{center}
	\begin{tabular}{C{1.5cm}||C{1.5cm}C{1.4cm}C{1.76cm}C{1.4cm}C{1.7cm}C{1.4cm}C{1.4cm}C{1.4cm}}\toprule[.8pt]
	            Vimeo90K-T &  bicubic &  TOFlow*~\cite{TOFlow-Vimeo} &
	  DUF-52L*~\cite{DUF} &
	  RBPN*~\cite{RBPN} &
	  EDVR-L*~\cite{EDVR} &
	  TGA~\cite{TGA} &
	  RSDN~\cite{RSDN} &
	  FDAN(Ours) \\ \midrule[.5pt]
	FLOPs(T)  &
	  - &
	  2.17 &
	  1.65 &
	  24.81 &
	  0.33 &
	  0.07 &
	  0.35 &
	  0.25 \\
	\#Param.(M) &
	  - &
	  1.41 &
	  5.82 &
	  12.2 &
	  20.7 &
	  5.8  &
	  6.19 &
	  8.97 \\
	Y &
	  31.17/0.8665 &
	  34.62/0.9212 &
	  36.87/0.9447 &
	  37.20/0.9458 &
	  {\textcolor{blue} {37.61}/0.9489} &
	  37.59/\textcolor{blue} {0.9516} &
	  37.05/0.9454 &
		{\textcolor{red} {37.75}/\textcolor{red} {0.9522}}	   \\
	RGB &
	  29.63/0.8460 &
	  32.78/0.9040 &
	  34.96/0.9313 &
	  35.39/0.9340 &
	  {\textcolor{blue} {35.79}/\textcolor{blue} {0.9374}} &
	  35.57/0.9387 &
	  35.14/0.9325 &
		{\textcolor{red} {35.91}/\textcolor{red} {0.9412}}
	   \\ \bottomrule[.8pt]
	\end{tabular}

	\begin{tabular}{C{1.5cm}||C{1.5cm}C{1.4cm}C{1.76cm}C{1.4cm}C{1.7cm}C{1.4cm}C{1.4cm}C{1.4cm}}\toprule[.8pt]
	UDM10 &
	  bicubic &
	  TOFlow*~\cite{TOFlow-Vimeo} &
	  DUF-52L*~\cite{DUF} &
	  RBPN*~\cite{RBPN} &
	  EDVR-L~\cite{EDVR} &
	  PFNL*~\cite{PFNL-UDM10} &
	  RSDN~\cite{RSDN} &
	  FDAN(Ours) \\ \midrule[.5pt]
	Y &
	  31.99/0.8950 &
	  36.26/0.9438 &
	  38.48/0.9605 &
	  38.66/0.9596 &
	  {\textcolor{blue} {39.44}/0.9646} &
	  38.74/0.9627 &
	  {39.35/\textcolor{blue} {0.9653}} &
	  {\textcolor{red} {39.91}/\textcolor{red} {0.9686}}
	  \\
	RGB &
	  30.57/0.8771 &
	  34.46/0.9298 &
	  36.78/0.9514 &
	  36.53/0.9462 &
	  37.27/0.9522 &
	  36.78/0.9514 &
	  {\textcolor{blue} {37.46}/\textcolor{blue} {0.9557}} &
	  {\textcolor{red} {37.68}/\textcolor{red} {0.9568}}
	  \\ \bottomrule[.8pt]
	\end{tabular}

	\begin{tabular}{C{1.5cm}||C{1.5cm}C{1.4cm}C{1.76cm}C{1.4cm}C{1.7cm}C{1.4cm}C{1.4cm}C{1.4cm}}\toprule[.8pt]
	Vid4 &
	  bicubic &
	  TOFlow*~\cite{TOFlow-Vimeo} &
	  DUF-52L*~\cite{DUF} &
	  RBPN*~\cite{RBPN} &
	  EDVR-L*~\cite{EDVR} &
	  TGA~\cite{TGA} &
	  RSDN~\cite{RSDN} &
	  FDAN(Ours) \\ \midrule[.5pt]
	Y &
	  23.58/0.6270 &
	  25.85/0.7659 &
	  27.38/0.8329 &
	  27.17/0.8205 &
	  27.35/0.8264 &
	  27.63/0.8423 &
	  {\textcolor{red} {27.92}/\textcolor{blue} {0.8505}} &
	  {\textcolor{blue} {27.88}/\textcolor{red} {0.8508}} 	  \\
	RGB &
	  22.17/0.6020 &
	  24.39/0.7438 &
	  25.91/0.8166 &
	  25.65/0.7997 &
	  25.83/0.8077 &
	  26.14/0.8258 &
	  {\textcolor{red} {26.43}/\textcolor{red} {0.8349}} &
	  {\textcolor{blue} {26.34}/\textcolor{blue} {0.8338}}	  \\ \bottomrule[.8pt]
	\end{tabular}
	\end{center}
	\vspace{.0cm}
	\end{table*}
	\begin{figure*}[] 

	\vspace{-.10cm}
	\setlength{\abovecaptionskip}{0cm}
	\setlength{\belowcaptionskip}{0pt}
	\centering
	  \subfigure[Vimeo-90K-T]{
	    \label{fig:subfig:6a} 
	    \includegraphics[scale=.65]{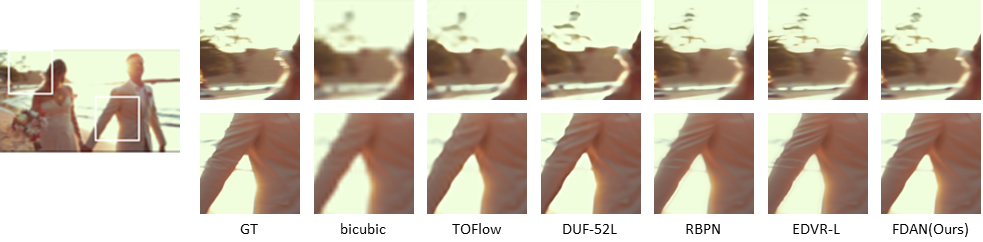}}
	  \hspace{1in}
	  \subfigure[UDM10]{
	    \label{fig:subfig:6b} 
	    \includegraphics[scale=.65]{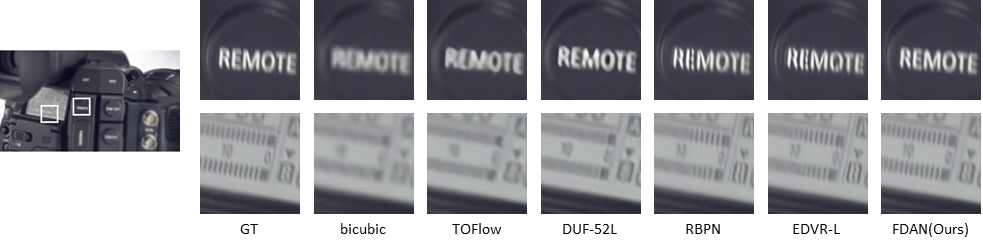}}
	  \hspace{1in}
	  \subfigure[Vid4]{
	    \label{fig:subfig:6c} 
	    \includegraphics[scale=.65]{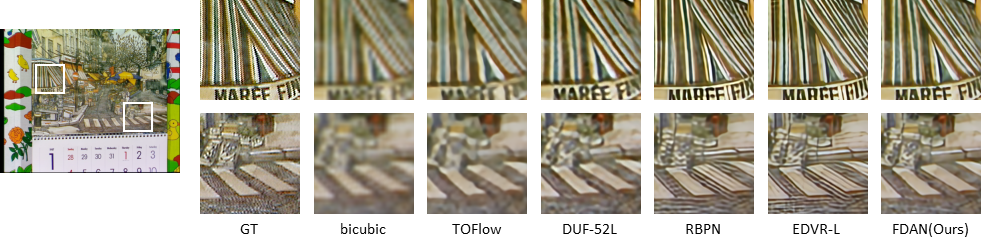}}

	\caption{Qualitative comparison on Vimeo-90K-T, UDM10 and Vid4 respectively for $4\times$ VSR. The patches are of size $100\times100$. Zoom in for better visualization.}
	\label{fig:SOTA}

	\vspace{-.20cm}
	\end{figure*}

\subsection{Comparison with State-of-the-arts}

	We compare our proposed method with seven state-of-the-art VSR approaches, containing TOFlow~\cite{TOFlow-Vimeo}, RBPN~\cite{RBPN}, EDVR~\cite{EDVR}, DUF~\cite{DUF}, PFNL~\cite{PFNL-UDM10}, TGA~\cite{TGA} and RSDN~\cite{RSDN}. In explicit alignment methods, TOFlow and RBPN use flow-base alignment while EDVR use deformable alignment. For implicit alignment methods like DUF, PFNL, TGA and RSDN . Tab.~\ref{table:1} shows quantitative results, where the number of parameters and FLOPs are also provided. Fig.~\ref{fig:SOTA} shows qualitative comparisons on Vimeo90K-T, UDM10 and Vid4. Note that EDVR uses bicubic degradation while others use bicubic and blur degradation as in our method.

	Our method achieves the state-of-the-art performance on Vimeo90K-T and UDM10, which contains various motion. In Vid4 with relatively slow motion and more complex textures, our method outperforms most of the methods. Therefore, it is validated that our method is capable of handling fast motion and generalize well to other datasets. Furthermore, our method consumes relatively less parameters and FLOPs in methods which employ explicit alignment. Note that TGA estimates homography between frames optionally to deal with fast motion so no parameters or FLOPs for pre-alignment is included in TGA, but may limit the inter-frame transformation to homography transformation.

	However, our method is insufficient to handle data with complex texture according to the quantitative comparison on Vid4. This may be due to naive design of the reconstruction module, which can be improved by applying advanced modules like residual dense block~\cite{RDN,DUF} or SD Block in RSDN~\cite{RSDN}, as well as other effective modules in SISR. Besides, as shown in the qualitative results, though our method cannot create sharp texture as RBPN or EDVR, it does not create artifacts and is consistent with the ground truth.


\subsection{Ablation Study}\label{subsection:Abla}
In this section, we verify the superiority of our alignment method FDA, including the introduction of flow to deformable alignment using MFE, and the integration strategy of flow to deformable convolution. We perform experiments with different settings in the alignment modules illustrated in Tab.~\ref{table:2}. Baseline uses 2 cascading deformable convolutional layers. M-2 has one more step than Baseline that applies MFE to warp neigboring feature maps first. M-3 is our final FDAN, applying MFE and FDC followed by a deformable convolutional layer (Fig.~\ref{fig:subfig:2}). M-1 applies the alignment module in EDVR~\cite{EDVR}, which employs 4 deformable convolutional layers (L1, L2, L3 and cascading) in a pyramid structure called PCD (Fig.~\ref{fig:subfig:1}) to solve fast motion. The number of parameters for each experiment including the one for alignment and the total one is also shown in Tab.~\ref{table:2}. Note that the alignment parameters include those for generating initial features in L2 and L3 in PCD, and those for MFE in flow-guided methods.

	\begin{table}[]
	\setlength{\abovecaptionskip}{0cm}
	\setlength{\belowcaptionskip}{-.0cm}
	\caption{Settings for experiments in ablation study. ``Align'' denotes the alignment methods. ``Flow. Inte'' denotes the integration strategy of flow to deformable alignment. }\label{table:2}
	\footnotesize

	\vspace{-.2cm}
	\begin{center}

	\begin{tabular}{c||cc|c}
	\toprule[.8pt]
	\begin{tabular}[c]{@{}l@{}}Experiment\end{tabular} &
	  \multicolumn{1}{c}{Align} &
	  \begin{tabular}[c]{@{}l@{}}Flow\\ Inte.\end{tabular} &
	  \begin{tabular}[c]{@{}c@{}}\#Param.(M)\\ Align / Total\end{tabular} \\ \midrule[.5pt]

	Baseline 	& 2dcnv     & -   	 	& 1.67 /  7.44  \\
	M-1 		& PCD   	& -   		& 5.12 / 10.89 \\
	M-2 		& naive FDA & Naive   	& 3.20 /  8.97  \\
	M-3 		& FDA 		& Advanced 	& 3.20 /  8.97  \\
	\bottomrule[.8pt]
	\end{tabular}

	\vspace{-.8cm}
	\end{center}
	\end{table}
	
	\begin{figure}[] 
	\setlength{\abovecaptionskip}{-.0cm}
	\setlength{\belowcaptionskip}{-.2cm}
	\centering
	 \subfigure[FDA]{\label{fig:subfig:2}\includegraphics[scale=0.78]{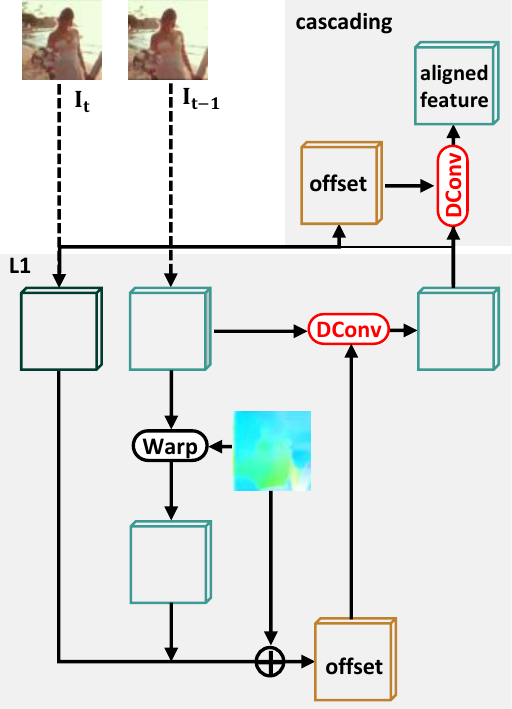}}
	 \hspace{0.02cm}
	 \subfigure[PCD]{\label{fig:subfig:1}\includegraphics[scale=0.78]{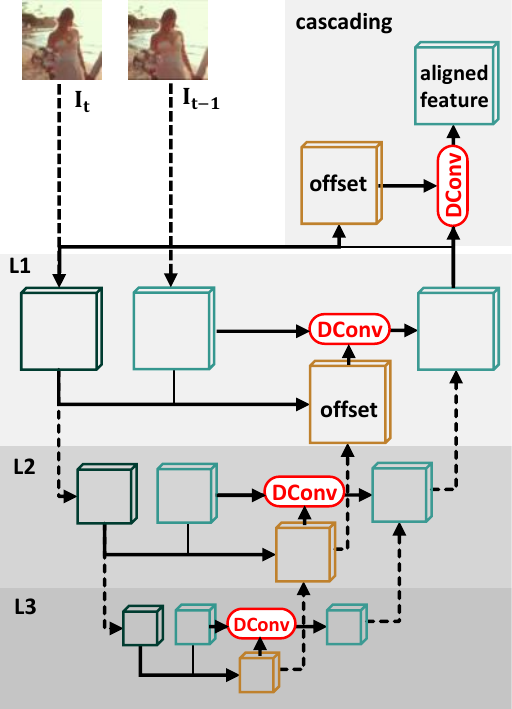}}
	\caption{A comparison of the alignment modules: FDA in M-3 (FDAN) and PCD in M-1.}\label{fig:1}
	\end{figure}

{\bf The introduction of flow to deformable alignment using MFE.}
We compare Baseline, M-1 and M-2 to verify the effectiveness of introducing flow. As PCD in M-1 applies a deformable alignment method, which has shown remarkable performance in handling videos with various motion~\cite{EDVR}, we compare it with our flow-guided deformable alignment method in M-2.

As shown in Tab~\ref{table:3}, M-2 has noticeable gains especially in datasets with fast motion compared to Baseline, which shows the effectiveness of introducing flow to deformable alignment. Comparing M-1 to M-2, in Vimeo90K-T that shares the same domain with the training dataset, M-1 is better than M-2 on videos with slow and medium motion, but even worse on videos with fast motion. While on the other three datasets from external domain, M-1 does not gain superiority as in Vimeo90K-T, and even performs worse on Vid4 with slow motion.

Furthermore, according to the visual result in Fig.~\ref{fig:vis_dcnv_1}, Baseline is unable to sample on the true corresponding locations. M-1 suffers from the same situation when the motion range is beyond its sampling range, indicating that it may only enlarge the sampling range compared to Baseline.
On the contrary, M-2 samples around the true corresponding locations, which reveals its superiority in handling fast motion. Therefore, naive FDA is competitive in handling fast motion and generalization ability, though with 15\% less FLOPs (on Vimeo90K-T) and 40\% less parameters than PCD.
	\begin{table*}[t]
	\setlength{\abovecaptionskip}{0cm}
	\setlength{\belowcaptionskip}{-.0cm}
	\caption{PSNR (dB) / SSIM comparison among different alignment methods for $4\times$ VSR. Vimeo90K-T is divided by motion velocity, ``total'' means the whole dataset. FLOPs is calculated on Vimeo90K-T whose HR images is of size $448\times 256$. \textcolor{red} {Red }text denotes the best performance.}\label{table:3}
	\footnotesize
	\begin{center}

	\begin{tabular}{C{1.1cm}||C{1.5cm}C{1.5cm}C{1.5cm}C{1.5cm}C{1.5cm}C{1.5cm}C{1.5cm}|C{.7cm}C{.8cm}}

	\toprule[.8pt]
	\multicolumn{1}{l||}{} &
	  \multicolumn{4}{c}{Vimeo90K-T} &
	   &
	   &
	   &
	   &
	   \\ \cline{2-5}
	\multicolumn{1}{l||}{\multirow{-2}{*}{\begin{tabular}[c]{@{}c@{}}Expe-\\ riment\end{tabular}}} &

	  fast &
	  medium &
	  slow &
	  total &
	  \multirow{-2}{*}{SPMC} &
	  \multirow{-2}{*}{UDM10} &
	  \multirow{-2}{*}{Vid4} &
	  \multirow{-2}{*}{\begin{tabular}[c]{@{}c@{}}FLOPs\\(T)\end{tabular}} &
	  \multirow{-2}{*}{\begin{tabular}[c]{@{}c@{}}FLOPs\\ \_4K(T)\end{tabular}} \\
	  \midrule[.5pt]
	Baseline &
		38.33/0.9537 &
		35.95/0.9437 &
		32.78/0.9130 &
		35.66/0.9389 &
		30.27/0.8785 &
		37.35/0.9545 &
		26.17/0.8288 &
	  0.19 &
	  14.52 \\
	M-1 &
	  38.76/0.9572 &
	  \textcolor{red} {36.21}/\textcolor{red} {0.9459} &
	  \textcolor{red} {32.94}/\textcolor{red} {0.9151} &
	  \textcolor{red} {35.93}/\textcolor{red} {0.9413} &
	  30.43/0.8803 &
	  37.62/0.9564 &
	  26.25/0.8317  &
	  0.27 &
	  20.59 \\	
	M-2 &
	  38.80/0.9579 &
	  36.11/0.9453 &
	  32.83/0.9138 &
	  35.85/0.9408 &
	  30.41/0.8816 &
	  37.61/0.9561 &
	  26.29/0.8326 &
	  0.23 &
	  18.79 \\
	M-3 &
	  \textcolor{red} {38.87}/\textcolor{red} {0.9584} &
	  36.17/0.9458 &
	  32.86/0.9141 &
	  35.91/0.9412 &
	  \textcolor{red} {30.45}/\textcolor{red} {0.8819} &
	  \textcolor{red} {37.68}/\textcolor{red} {0.9568}  &
	  \textcolor{red} {26.34}/\textcolor{red} {0.8338}  &
	  0.23 &
	  18.79 \\

	  \bottomrule[.8pt]

	\end{tabular}

	\end{center}

	\end{table*}

	\begin{figure}[] 
	\setlength{\abovecaptionskip}{.4cm}
	\setlength{\belowcaptionskip}{-.6cm}
	\centering
	\includegraphics[width=3.3in]{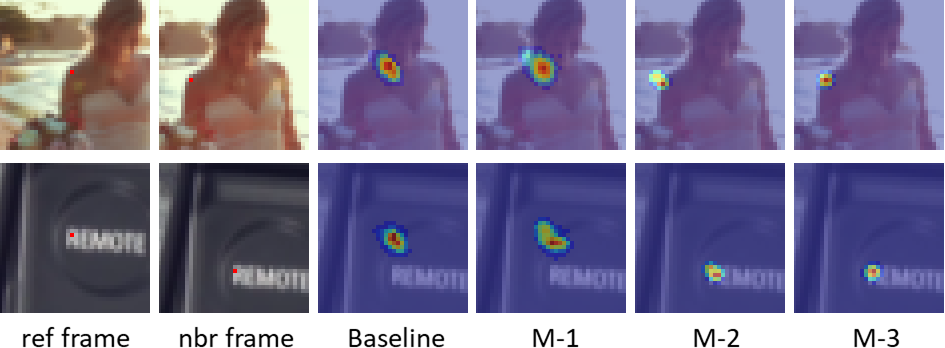}

	\caption{The visualization of sampling on the neighboring frames. The red point in the reference frame is the location to compensate and the one in the neighboring frame is the corresponding location obtained from the flow from M-3. Comparsion of Baseline, M-1 and M-2 shows the effectiveness of introducing flow to deformable alignment. Comparsion of M-2 and M-3 shows the effectiveness of Advanced Integration. }\label{fig:vis_dcnv_1}
	\end{figure}

	\begin{figure*}[] 
	\setlength{\abovecaptionskip}{-.2cm}
	\setlength{\belowcaptionskip}{-.5cm}
	\begin{center}
    \vspace{-.2cm}
	\includegraphics[scale=.32]{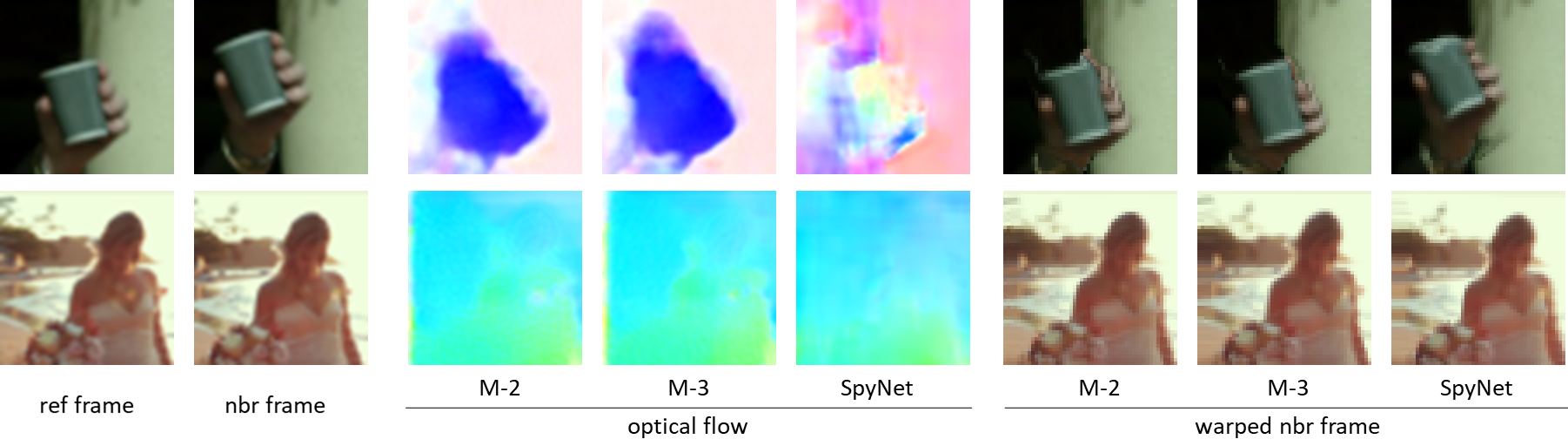}
	\end{center}

	\caption{The visualization of the generated flow and warped neighboring frames using Naive Integration, Advanced Integration and SpyNet. }\label{fig:abla_offset}
	\end{figure*}

{\bf The integration strategy of flow to deformable alignment.} We compare Naive Integration and Advanced Integration to verify the superiority of Advanced Integration. As the quantitative result shown in Tab.~\ref{table:3}, Advanced Integration outperforms Naive Integration over all the datasets without extra FLOPs or parameter consumption. For visualization results, Advanced Integration can perform more precise sampling and generate better optical flow. Fig.~\ref{fig:vis_dcnv_1} shows the visualization of the sampling locations on the L1 deformable convolutional layer. Compared to M-2, M-3 involves less noises by, for example, suppressing sampling on background when compenstating foreground. Fig.~\ref{fig:abla_offset} shows the generated flow and the corresponding warped neighboring frames. We also show the results calculated by pretrained SpyNet~\cite{SpyNet} for reference, which is widely used in VSR~\cite{TOFlow-Vimeo, TDAN, BasicVSR}. The edges of the flow in M-3 is clearer than those in M-2, while SpyNet cannot capture the fast motion well especially those on small objects of degraded images.

{\bf Discussion.}
\emph{(1) Why does FDA perform worse than PCD in slow and medium motion in Vimeo90K-T?} 
In FDA, flow is introduced to deal with various motion especially fast one. Assuming there were no motion between frames, then FDA would degrade to the alignment module in Baseline, while PCD would be a multi-scale feature encoder with over twice more parameters. So PCD may benefits from multi-scale feature extraction and an abundance of parameters. However, on the other hand, PCD may overfit Vimeo90K as it performs worse than FDA in external datasets, even in Vid4 with slow motion.
\emph{(2) Why can FDA handle fast motion better than PCD?} 
In MFE, we perform all-pairs matching to search correspondences globally. While PCD can only find the local optimum by regressing to the ground truth HR, which does not directly supervise the alignment module to find the true correspondence especially in extreme scenes such as fast motion.
We can also discuss this issue in the view of the offset generation process in deformable convolution. As Eq.(7) in ~\cite{DeformableCN}, during the back propagation step, the gradient of offset in a specific spatial location only depends on its spatial neighbors with radius=1, so the global optimum sampling location cannot be observed as the feature maps is usually non-convex. With the pyramid design in PCD, this problem is alleviated when searching in feature maps with lower resolution, but cannot be totally solved. As shown in ~\ref{fig:offsetDistri_fast}, the value of offsets are mostly small for fast motion. Furthermore, the perceptive field of the offset generation module using only three or four convolutional layers is limited, which also makes fast motion difficult to capture.
\emph{(3) Why can FDA generalize better than PCD?} %
In FDA, applying the matching score ranking mechanism within the target dataset can suffer less from overfitting on the training datasets. Furthermore, the features to perform matching is generated by pre-trained Resnest50 that focus more on semantic representation with better generalization ability~\cite{SFNet}. While in PCD, the feature extraction may focus more on capturing low-level features like texture thus has limited generalization ability.
\section{Conclusion}
In this paper, we introduce optical flow to deformable alignment for VSR to handle fast motion and further explore the integration strategy of flow and deformable convolution. Specifically, we propose an Flow-guided Deformable Alignment method to help sample on neighboring frames precisely and make full use of temporal compensatory information. The proposed Flow-guided Deformable Alignment Network (FDAN) is capable of dealing with fast motion robustly with modest computation and memory consumption and reaches state-of-the-art on several benchmark datasets.

{\small
\bibliographystyle{ieee_fullname}
\bibliography{myegbib}
}

\end{document}